\pdfoutput=1
\documentclass[11pt]{article}
\usepackage{nccmath}
\usepackage{graphicx}
\usepackage[]{acl}
\usepackage{amsmath,amsfonts,bm}

\def\eqref#1{equation~\ref{#1}}
\def\1{\bm{1}}

\DeclareMathAlphabet{\mathsfit}{\encodingdefault}{\sfdefault}{m}{sl}
\SetMathAlphabet{\mathsfit}{bold}{\encodingdefault}{\sfdefault}{bx}{n}

\usepackage{url}
\usepackage{times}
\usepackage{latexsym}
\usepackage{enumitem}
\usepackage{booktabs}
\usepackage{multirow}
\usepackage{colortbl}
\usepackage{geometry}
\usepackage{float}
\usepackage{array}
\usepackage{subfig}
\usepackage[T1]{fontenc}
\usepackage[utf8]{inputenc}

\usepackage{microtype}

\title{Uncertainty Quantification for In-Context Learning of Large Language Models}

\author{
    Chen Ling\textsuperscript{1}, Xujiang Zhao\textsuperscript{2}, \textbf{Xuchao Zhang}\textsuperscript{3}, Wei Cheng\textsuperscript{2}, Yanchi Liu\textsuperscript{2}, \\\textbf{Yiyou Sun}\textsuperscript{2}, \textbf{Mika Oishi}\textsuperscript{4},
\textbf{Takao Osaki}\textsuperscript{4}, \textbf{Katsushi Matsuda}\textsuperscript{4}, \\\textbf{Jie Ji}\textsuperscript{1}, \textbf{Guangji Bai}\textsuperscript{1}, \textbf{Liang Zhao}\textsuperscript{1}, \textbf{Haifeng Chen}\textsuperscript{3} \\
    \textsuperscript{1}Emory University,
    \textsuperscript{2}NEC Labs America, \textsuperscript{3}Microsoft,
    \textsuperscript{4}NEC Corporation\\
    \texttt{chen.ling@emory.edu, xuzhao@nec-labs.com, liang.zhao@emory.edu}
}

\begin{document}
\maketitle
\begin{abstract}
% Uncertainty estimation is important for ensuring the safety and robustness of AI systems. While most research in the area has focused on unstructured prediction tasks, limited work has investigated general uncertainty estimation approaches for structured prediction. 
% In this work, we propose two forms of uncertainty decomposition from the Bayesian perspective for white-box LLMs and black-box LLMs, respectively.
In-context learning has emerged as a groundbreaking ability of Large Language Models (LLMs) and revolutionized various fields by providing a few task-relevant demonstrations in the prompt. However, trustworthy issues with LLM's response, such as hallucination, have also been actively discussed. Existing works have been devoted to quantifying the uncertainty in LLM's response, but they often overlook the complex nature of LLMs and the uniqueness of in-context learning. In this work, we delve into the predictive uncertainty of LLMs associated with in-context learning, highlighting that such uncertainties may stem from both the provided demonstrations (aleatoric uncertainty) and ambiguities tied to the model's configurations (epistemic uncertainty). We propose a novel formulation and corresponding estimation method to quantify both types of uncertainties. The proposed method offers an unsupervised way to understand the prediction of in-context learning in a plug-and-play fashion. Extensive experiments are conducted to demonstrate the effectiveness of the decomposition. The code and data are available at: \url{https://github.com/lingchen0331/UQ_ICL}.

%Large Language Models (LLMs) have significantly impacted various fields with their capacity for in-context learning, enabling rapid adaptation to new tasks with few-shot demonstrations. However, with the increasing reliance on LLMs, understanding and quantifying the uncertainty associated with their predictions becomes paramount. This paper delves into the predictive uncertainty associated with the in-context learning of LLMs, highlighting that such uncertainties may stem from both the provided demonstrations (aleatoric uncertainty) and ambiguities tied to the model's configurations (epistemic uncertainty). Compared to existing works that only offer a generalized measure of uncertainty, our research pioneers a comprehensive decomposition of uncertainty in LLMs' in-context learning, categorizing them into their primary aleatoric and epistemic sources. The proposed method offers a novel and unsupervised way to understand the prediction of LLMs without modifying their architectures in a plug-and-play fashion. Extensive experiments are conducted to demonstrate the effectiveness of the decomposition. The code and data are available at \url{anonymous.4open.science/r/UQ_ICL-6BF5}.
\end{abstract}

\section{Introduction}
Large Language Models (LLMs) have revolutionized diverse domains by serving as general task solvers, which can be largely attributed to the emerging capability: \textit{in-context learning}. By providing demonstrations of the task to LLMs as part of the prompt, LLMs can quickly grasp the intention and make corresponding responses to the particular task \citep{min2022rethinking}. In this paradigm, LLMs can quickly adapt to solve new tasks at inference time (without any changes to their weights). Advanced LLMs, e.g., GPT-4 and LLaMA, have achieved state-of-the-art results on LAMBADA (commonsense sentence completion), TriviaQA (question answering) \citep{xie2021explanation}, and many tasks in other domains \citep{ling2023domain,ling2023open}. 

While in-context learning has achieved notable success, LLMs remain vulnerable to well-known reliability issues like hallucination \cite{rawte2023survey,bai2024beyond}. Uncertainty quantification has emerged as a popular strategy to assess the reliability of LLM responses. In the past two years, numerous works \citep{xiao2022uncertainty, lin2023generating, ling2023improving, amayuelas2023knowledge, kuhn2023semantic} have been proposed to quantify the uncertainty of LLM response. These approaches could return a confidence score or directly compute variance/entropy across multiple LLM responses; however, they often overlook the complex nature of LLMs and their reliance on provided demonstrations in in-context learning, so that existing methods may not provide insights into the underlying causes or the interactions among different factors contributing to uncertainty.

%Despite the success of in-context learning, LLMs are also proven to be vulnerable due to wide-known trustworthy issues like hallucination. Uncertainty quantification is one of the most popular approaches to answer whether we can trust LLM's response. In the past two years since the introduction of LLMs, many works \citep{xiao2022uncertainty,lin2023generating,ling2023improving,amayuelas2023knowledge,kuhn2023semantic} have been proposed to quantify the uncertainty of LLM's response by 1) either training a surrogate model or directly let LLM itself to return a confidence score; or 2) empirically quantifying the uncertainty of LLM's outputs by calculating their variance/entropy of multiple responses. These methods can give a measure of uncertainty, but they may not elucidate the underlying causes or the interactions between different factors causing the uncertainty. 

\begin{figure*}[t]
\centering
\includegraphics[width=0.9\textwidth]{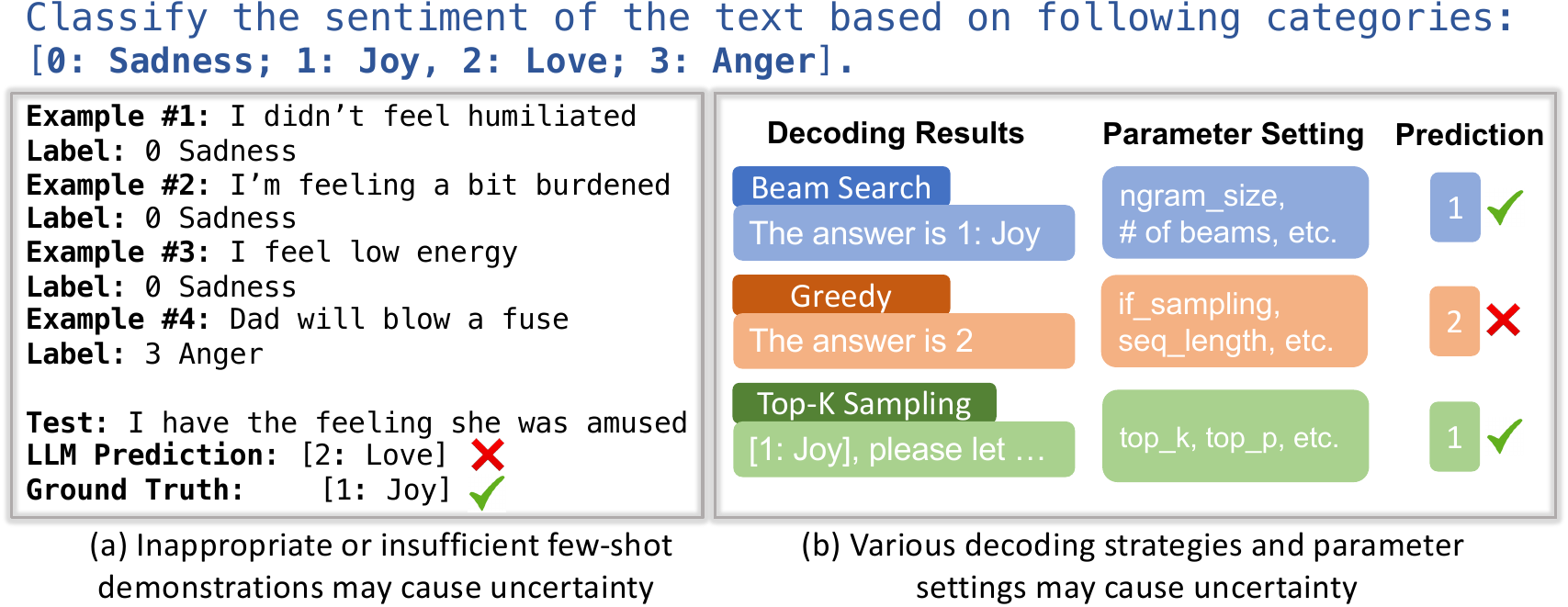}
\vspace{-3mm}
\caption{Uncertainty in LLM's prediction can stem from two aspects: a) \textit{Demonstration Quality}: LLMs are likely to make wrong predictions if the demonstrations are inappropriate; b) \textit{Model Configuration}: different decoding strategies (e.g., {beam\_search} and {top\_k sampling}) and their parameter settings may return different predictions.}
\label{fig: intro}
\vspace{-5mm} 
\end{figure*}

A natural question therefore arises: when LLM uses in-context learning to predict a wrong answer with high uncertainty, can we indicate if it is caused by the demonstration examples or by the model itself? Given LLM's responses to a particular task, it's essential to decompose the uncertainty into its primary sources to address the question. Specifically, \textit{Aleatoric Uncertainty (AU)} refers to variations in the data, often linked to the demonstration examples. As shown in Figure \ref{fig: intro} (a), LLM's output can easily be disturbed by inappropriate demonstrations since the provided demonstrations do not cover all possible labels. The noise and potential ambiguity of these demonstrations could bring uncertainty, which, in turn, may hinder the accuracy of the response. In contrast, \textit{Epistemic Uncertainty (EU)} stems from ambiguities related to the model parameters or different configurations. As depicted in Figure \ref{fig: intro} (b), different decoding strategies (e.g., beam search and greedy decoding) and their hyperparameter settings can have different decoding results. Recognizing and quantifying the uncertainty from the model's perspective can also be critical in evaluating the generated responses, which allows us to understand the model's confidence level toward the task and make necessary adjustments (e.g., choosing a more powerful model or conducting an ensemble prediction).

Despite the strides made by existing works in understanding the total uncertainty, the decomposition of uncertainty in the realm of in-context learning remains under-explored. In this work, we propose a novel framework for quantifying the uncertainty of in-context learning to aleatoric and epistemic components from the generated outputs. Our contributions are summarized as follows.
\begin{itemize}[leftmargin=*]
\itemsep0em
    \item \textbf{Problem.} We formulate the problem of uncertainty decomposition that extracts epistemic and aleatoric uncertainties from the predictive distribution of LLMs with in-context learning.
    \item \textbf{Method.} We propose quantifying both aleatoric and epistemic uncertainty from the mutual information perspective. A novel entropy-based estimation method is also designed to handle the free-form outputs of LLMs.
    \item \textbf{Experiment.} Extensive experiments are conducted to evaluate different aspects of uncertainty, followed by specific applications and case studies to show how two types of uncertainty influence the model's performance. 
\end{itemize}

\section{Uncertainty Decomposition of In-context Learning}
We first formulate the process of in-context learning as Bayesian Neural Networks with latent variables. Based on the formulation, we propose to decompose the predictive uncertainty into its epistemic and aleatoric components from the mutual information perspective, followed by a novel way to estimate both uncertainties based on the entropy of the prediction's distribution.

\subsection{Background}
LLMs are typically trained using maximum likelihood estimation on a large corpus of text. The training goal is to maximize the likelihood of the observed data under the model: $\mathcal{L}(\Theta) = \prod_{i\le N} p(\omega_i | \omega_1, \omega_2, \ldots, \omega_{i-1}; \Theta)$, where each $\omega_i\in \mathbf{x}$ is a token in a sentence $\mathbf{x}=[\omega_1, \ldots, \omega_N]$, and $\Theta$ denotes the set of parameters. 

\paragraph{Latent Concept.} From the Bayesian point of view, LLM's in-context learning ability is obtained by mapping the training token sequence $\mathbf{x}$ to a latent \textit{concept} $z$ \cite{xie2021explanation}. The concept $z$ is a latent variable sampled from a space of concepts $\mathcal{Z}$, which defines
a distribution over observed tokens $\omega_i$ from a training context $\mathbf{x}$:
\begin{equation*}
    p(\omega_1, \ldots, \omega_N)=\int_{z\in \mathcal{Z}}p(\omega_1, \ldots, \omega_N|z)p(z)dz.
\end{equation*}
The concept can be interpreted as various document-level statistics, such as the general subject matter of the text, the structure/complexity of the text, the overall emotional tone of the text, etc. 

\paragraph{In-context Learning.} 
Given a list of independent and identically distributed (i.i.d.) in-context demonstrations (contain both question and answer) $[\mathbf{x}_1, \ldots, \mathbf{x}_{T-1}]$ concatenated with a test question (without the task answer) $\mathbf{x}_{T}$ as prompt. Each demonstration $\mathbf{x}_i$ in the prompt is drawn as a sequence conditioned on the same concept $z$ and describes the task to be learned. LLMs generate a response $\mathbf{y}_T$ to the test question $\mathbf{x}_{T}$ based on the aggregated prompt $\mathbf{x}_{1:T}$:
\begin{align}
p(\mathbf{y}_T|\mathbf{x}_{1:T})=\int_{z\in\mathcal{Z}}p(\mathbf{y}_T|\mathbf{x}_{1:T},z)p(z|\mathbf{x}_{1:T})dz.\nonumber
\end{align}
In-context learning can be interpreted as \textit{locating} a pre-existing concept $z$ based on the provided demonstrations $\mathbf{x}_{1:T-1}$, which is then employed to tackle a new task $\mathbf{x}_{T}$. Including more high-quality demonstrations within the prompt can help refine the focus on the relevant concept, enabling its selection through the marginalization term $p(z|\mathbf{x}_{1:T})$. Note that formulating in-context learning as Bayesian inference with latent variables is more of a hypothesis; however, demystifying the in-context learning from the view of Bayesian inference offers a probabilistic interpretation of how LLM learns and adapts to new data in context. 

In this work, we focus on quantifying the predictive uncertainty of LLMs in deterministic NLP tasks, such as text classification. Specifically, we address tasks where the training dataset $\mathcal{D}=\{\mathcal{X}, \mathcal{Y}\}$ consists of token sequences $\mathcal{X}=\{\mathbf{x}\}$ and their corresponding target outputs $\mathcal{Y}=\{\mathbf{y}\}$. For LLMs, the generation process is defined by the function $\mathbf{y}=f(\mathbf{x}, z; \Theta)$, where $f: \mathcal{X} \times \mathcal{Z}\rightarrow \mathcal{Y}$ is a deterministic function. The output $\mathbf{y}$ exhibits stochastic behavior, influenced by the latent concept $z\sim p(z|\mathbf{x}_{1:T})$ and the model parameters/configurations $\Theta$ (e.g., temperature, etc.). %Now, turning to the training objective, LLMs aim to approximate the exact posterior $p(\mathcal{W}, z| \mathcal{D})$ with the joint posterior $q(\mathcal{W}, z)$.

\subsection{Predictive Uncertainty Formulation of In-context Learning}
We formulate the predictive distribution of in-context learning for predicting $\mathbf{y}_{T}$ given few-shot demonstrations $\mathbf{x}_{1:T-1}$ and a test case $\mathbf{x}_{T}$ as:
\begin{align}\label{eq:BNN}
p(\mathbf{y}_{T}|\mathbf{x}_{1:T})\approx&\int p(\mathbf{y}_{T}|\Theta,\mathbf{x}_{1:T},z)\\&\cdot p(z|\mathbf{x}_{1:T})q(\Theta)dz\:d\Theta,\nonumber
\end{align}
where $p(\mathbf{y}_{T}|\Theta,\mathbf{x}_{1:T},z)$ is approximated by a Bayesian Neural Network-based likelihood function $\mathcal{N}(f(\mathbf{x}_{1:T}, z), \boldsymbol \Sigma)$, and $\boldsymbol\Sigma$ is the covariance matrix contains the variances and covariances associated with LLM parameters. $q(\Theta)$ is the approximated posterior of the LLM's parameters $\Theta$. Eq. (\ref{eq:BNN}) approximates LLM outputs following a Gaussian distribution, which serves as an initial framework for generating predictions based on input data and accompanying demonstrations: $p(\mathbf{y}_T|\mathbf{x}_{1:T})$, which entangles different types of uncertainties. We first present the overall pipeline of our uncertainty quantification framework, followed by formulation on decomposing the total uncertainty based on mutual information (Sec. \ref{sec: ent}) and a novel way to estimate the uncertainty (Sec. \ref{sec: est}). Note that LLMs can be categorized into white-box and black-box models \citep{ling2023domain} based on their transparency. Quantifying mutual information involves accessing the probability of generated tokens, which is not applicable to black-box LLMs. In this study, we also provide a decomposition way from the variance perspective for black-box LLMs. Due to the space limit, the variance-based decomposition can be found in Appendix \ref{sec: var}.

%White-box LLMs (e.g., LLaMA models) give access to parameter space and metadata of the generation. In contrast, black-box LLMs (e.g., GPT models) permit only the retrieval of generated responses. In this study, we present two methodologies for decomposing these uncertainties from the mutual information perspective for white-box LLMs (Sec. \ref{sec: ent}) and the variance perspective for black-box LLMs, respectively. Due to the space limit, the variance-based decomposition can be found in Appendix \ref{sec: var}. 

%As shown in Eq. (\ref{eq:BNN}), the uncertainty of the output $\mathbf{y}_{T}$ comes from the model parameter's perspective (i.e., epistemic uncertainty): $\Theta \sim q(\Theta)$ and input data's perspective (i.e., aleatoric uncertainty): $z\sim p(z|\mathbf{x}_{1:T})$ \cite{abdar2021review}. 

\begin{figure}[t]
\centering
\includegraphics[width=0.45\textwidth]{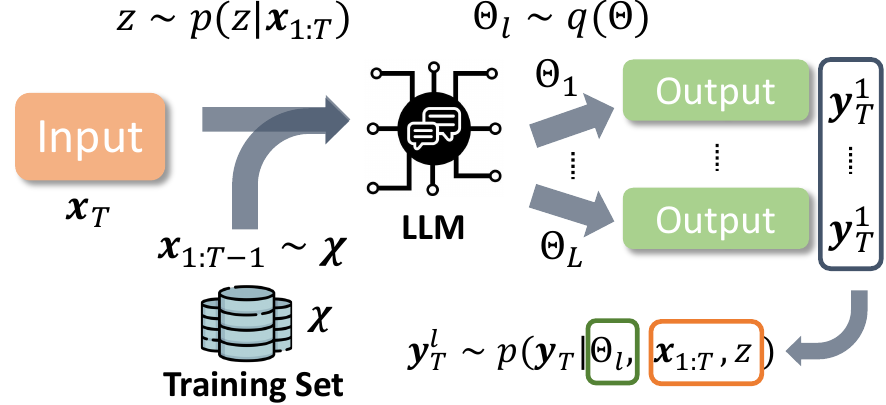}
\vspace{-3mm}
\caption{Uncertainty Quantification of In-context Learning Pipeline: we want to quantify the uncertainty that comes from 1) different in-context demonstrations $\mathbf{x}_{1:T}$; and 2) different model configurations $\Theta_l$.}
\label{fig: framework_simple}
\vspace{-5mm} 
\end{figure}

\paragraph{Framework Pipeline.} In this work, we employ a Bayesian framework to quantify the predictive uncertainty from LLMs, and the overall pipeline is visualized in Figure \ref{fig: framework_simple}. Specifically, the input $\mathbf{x}_{1:T}$ is composed of a test query $\mathbf{x}_{T}$ and a set of demonstrations $\mathbf{x}_{1:T-1}$ sampled from $\mathcal{X}$. By sampling different model parameters/configurations $\Theta_l \sim q(\Theta)$, LLM can return different outputs $\mathbf{y}^l_T\in[\mathbf{y}^1_T, \cdots, \mathbf{y}^L_T]$ based on the conditional probability $p(\mathbf{y}_{T}|\Theta_l,\mathbf{x}_{1:T},z)$. The collection of outputs $[\mathbf{y}^1_T, \cdots, \mathbf{y}^L_T]$ records the total uncertainty regarding $\Theta_l$ and demonstrations $\mathbf{x}_{1:T-1}$.
%We iterate the process several times with different $\mathbf{x}_{1:T-1}$, we can obtain different sets of responses corresponding to different $\mathbf{x}_{1:T-1}$. The variance of these responses comes from two uncertainty sources, i.e., 1) demonstrations $\mathbf{x}_{1:T-1}$; and 2) model parameters $\Theta$.

%In this study, we present two methodologies for decomposing these uncertainties from the mutual information perspective (Sec. \ref{sec: ent}) and variance perspective (Sec. \ref{sec: var}), catering to both white-box and black-box LLMs, respectively.

\begin{figure*}[t]
\centering
\includegraphics[width=0.89\textwidth]{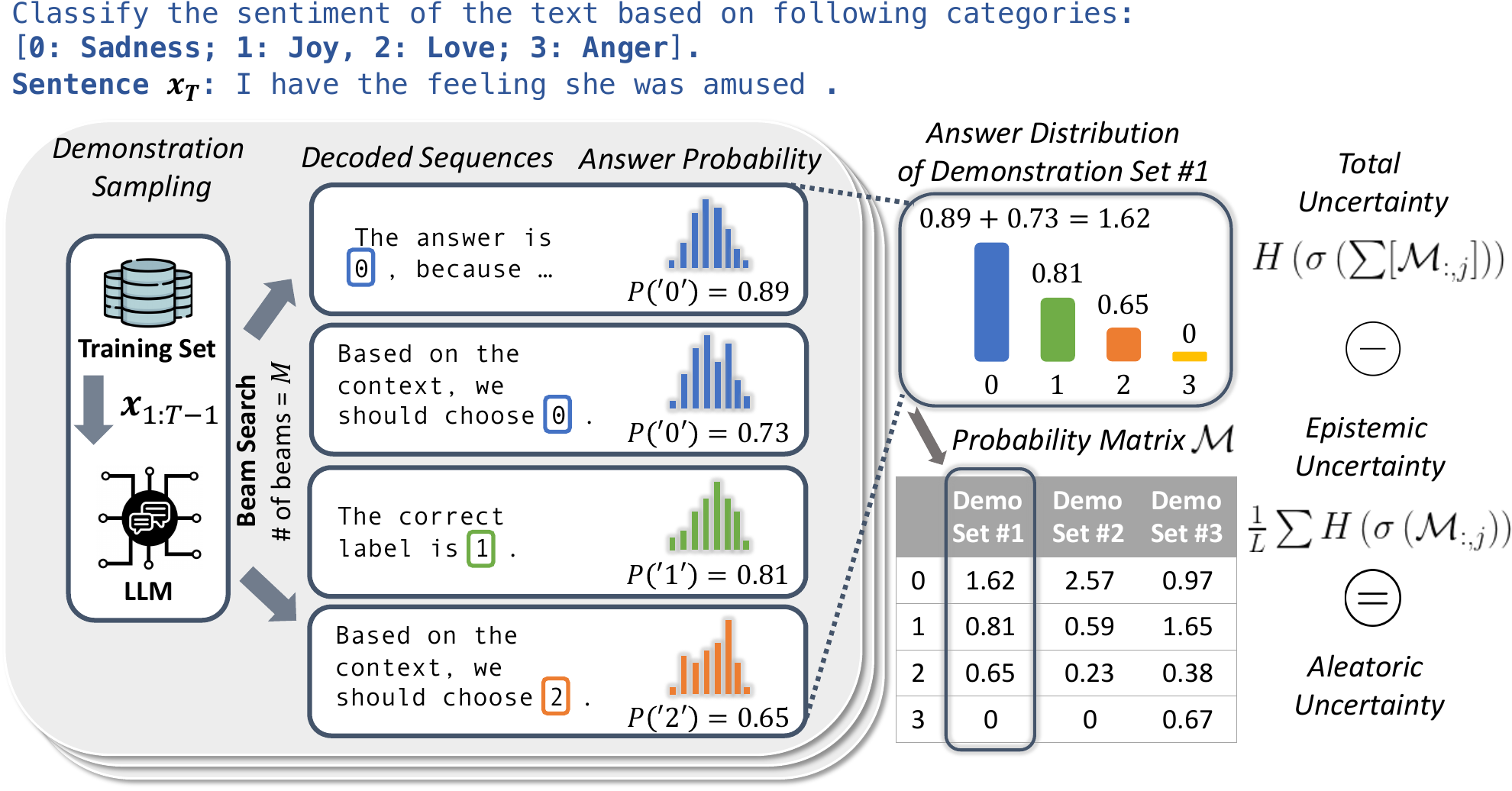}
\vspace{-1mm}
\caption{Framework of entropy-based uncertainty estimation, which consists of 1) generating $M$ sequences based on a set of $\mathbf{x}_{1:T-1}$; 2) selecting token(s) that is relevant to the answer and extract the probabilities; 3) aggregating the token probabilities of $M$ sequences into a distribution of predicted labels; 4) iterating the process $L$ times corresponding to $L$ different demonstration sets and form a probability matrix $\mathcal{M}$, where the column denotes different demonstration sets and the row denotes labels of the dataset.}
\label{fig: framework}
\vspace{-3mm} 
\end{figure*}

%We first decode multiple sequences based on a set of demonstrations randomly sampled from the training set. For each sequence, we normalize the score of each token into a probability distribution and we identify the token(s) that represents the answer. By aggregating the probability of each answer token(s) from each decoded sequence, the answer distribution can be obtained.

\subsection{Entropy-based Decomposition}\label{sec: ent}
As a widely adopted measure of uncertainty, entropy provides a quantifiable and interpretable metric to assess the degree of confidence in the model's predictions \citep{malinin2020uncertainty}. Since white-box LLMs can return the probability of each token in the generated sequence, it naturally makes entropy-based uncertainty measures applicable uniformly across different types of white-box LLMs.

\paragraph{Epistemic Uncertainty (EU).} Let $H(\cdot)$ be the differential entropy of a probability distribution, the total uncertainty in Eq. (\ref{eq:BNN}) can be quantified as $H\left(\mathbf{y}_T|\mathbf{x}_{1:T}\right)$, which entangles both aleatoric (i.e., demonstration $\mathbf{x}_{1:T-1}$) and epistemic (i.e., model parameter $\Theta$) uncertainties. To estimate the EU, we condition Eq. (\ref{eq:BNN}) on a specific realization of the model parameter $\Theta$, yielding $p(\mathbf{y}_T|\mathbf{x}_{1:T}, \Theta)=\int p(\mathbf{y}_T|\mathbf{x}_{1:T}, z, \Theta)p(z|\mathbf{x}_{1:T}) dz$ with an associated entropy $H(\mathbf{y}_T|\mathbf{x}_{1:T},z,\Theta)$. The expected value of this entropy under different demonstration sets can be expressed as $\mathbb{E}_{z}\left[H(\mathbf{y}_T|\mathbf{x}_{1:T},z,\Theta)\right]$, which serves as a metric to quantify the EU in Eq. (\ref{eq:BNN}). 

\paragraph{Aleatoric Uncertainty (AU).} In terms of AU, the randomness comes from different sets of demonstration $\mathbf{x}_{1:T-1}$ and their corresponding latent concept $z$. To estimate AU, we can quantify the mutual information between $\mathbf{y}_{T}$ and latent concept $z$, which can often be leveraged as an evaluation metric of AU \citep{wimmer2023quantifying}. As we have already obtained the EU, AU can subsequently be calculated as the discrepancy between the total uncertainty and the epistemic uncertainty: 
\begin{align}\label{eq:mutualinfo}
I(\mathbf{y}_{T}, z|\Theta) = & H\left(\mathbf{y}_{T}|\mathbf{x}_{1:T}, \Theta\right)\\&-\mathbb{E}_{z}\left[H(\mathbf{y}_T|\mathbf{x}_{1:T},z,\Theta)
%\mathbb{E}_{q(\Theta)}\left[H(\mathbf{y}_T|\mathbf{x}_{1:T},\Theta)
\right].\nonumber
\end{align}
%Let $H(\cdot)$ be the differential entropy of a probability distribution. The overall uncertainty in Eq. (\ref{eq:BNN}) can be quantified as $H\left(\mathbf{y}_T|\mathbf{x}_{1:T}\right)$. Generally, for LLMs, we typically have access to a deterministic set of parameters denoted by $\Theta$. As such, we do not marginalize $\Theta$ in Eq. (\ref{eq:BNN}). Instead, we condition the equation on a specific realization of this parameter set, yielding $p(\mathbf{y}_T|\mathbf{x}_{1:T}, \Theta)=\int p(\mathbf{y}_T|\mathbf{x}_{1:T}, z, \Theta)p(z|\mathbf{x}_{1:T}) dz$, with an associated entropy $H(\mathbf{y}_T|\mathbf{x}_{1:T},\Theta)$.  The expected value of this entropy under the distribution $q(\Theta)$ can be expressed as $\mathbb{E}_{q(\Theta)}\left[H(\mathbf{y}_T|\mathbf{x}_{1:T},\Theta)\right]$. This expectation serves as a metric to quantify the aleatoric uncertainty in Eq. (\ref{eq:BNN}) coming from $z$. In terms of epistemic uncertainty, the mutual information between $\mathbf{y}_{T}$ and $\Theta$ can often be leveraged as an evaluation metric \citep{wimmer2023quantifying}. In this case, the epistemic uncertainty in Eq. (\ref{eq:BNN}) can subsequently be calculated as the discrepancy between the total uncertainty and the aleatoric uncertainty:
% \begin{align}\label{eq:mutualinfo}
% I(\mathbf{y}_{T}, \Theta) = & H\left(\mathbf{y}_{T}|\mathbf{x}_{1:T}\right)\\&-\mathbb{E}_{q(\Theta)}\left[H(\mathbf{y}_T|\mathbf{x}_{1:T},\Theta)\right].\nonumber
% \end{align}
The entropy $H\left(\mathbf{y}_{T}|\mathbf{x}_{1:T},\Theta\right)$ can be approximately calculated as $-\sum_t \left[p(\omega^{\mathbf{y}_{T}}_t)\cdot\log p\left(\omega^{\mathbf{y}_T}_t\right)\right]$, where $p(\omega^{\mathbf{y}_T}_t)$ represents the probability of each possible next token $\omega^{\mathbf{y}_{T}}_t$ given the input prompt $\mathbf{x}_{1:T}$. Therefore, the AU in Eq. (\ref{eq:mutualinfo}) can be approximated by sampling many $z$ (by sampling different sets of demonstrations) to obtain different $\mathbf{y}_{T}$ conditioning on one set of parameters $\Theta$. The group of $\mathbf{y}_{T}$ can then be used to approximate the respective entropies for each group of demonstrations $\mathbf{x}_{1:T-1}$:
\begin{align}\label{eq: entropy_appro}
    &I(\mathbf{y}_{T},z|\Theta)\\\nonumber&= H\left(\mathbf{y}_{T}|\mathbf{x}_{1:T},\Theta\right)-\mathbb{E}_{z}\left[H(\mathbf{y}_{T}|\mathbf{x}_{1:T},z,\Theta)\right]\\&\approx \sum^{M\times L} H(\mathbf{y}_{T}) - \frac{1}{M}\sum^{M}_{m=1}\sum^{L}_{l=1}\left[H(\mathbf{y}_{T}^{\Theta_m, l})\right],\nonumber
\end{align}
where $[\mathbf{y}^{\Theta_m, l}_{T}]$ are obtained corresponding to different demonstrations $[\mathbf{x}^{1}_{1:T-1}, \ldots, \mathbf{x}^{L}_{1:T-1}]$, and $[\Theta_1, \ldots, \Theta_M]$ are sampled from $q(\Theta)$. However, in many cases, direct sampling from the posterior is hard since it requires a prohibitive number of samples to approximate it effectively. Beam search is then used as an efficient alternative to find high-quality hypotheses. This approach can be viewed as a form of importance sampling, where hypotheses are drawn from high-probability regions of the space. Each hypothesis $\mathbf{y}_T$ observed during the beam search process is associated with uncertainty, which is importance-weighted in proportion to $p(\mathbf{y}_T|\mathbf{x}_{1:T}, z)$. Beam Search thus serves as a practical and efficient way to sample from the posterior by focusing on the most relevant parts of the hypothesis space.
% For some LLMs that do not allow sampling different sets of parameters from the learned  $q(\Theta)$ as a standard Bayesian Neural Network, we can instead leverage Beam Search to enable stochastic output from LLMs. Direct sampling from the posterior is hard since it requires a prohibitive number of samples to approximate it effectively. Beam search is then used as an efficient alternative to find high-quality hypotheses. This approach can be viewed as a form of importance sampling, where hypotheses are drawn from high-probability regions of the space. Each hypothesis 
%  observed during the beam search process is associated with uncertainty, which is importance-weighted in proportion to 
% . Beam Search thus serves as a practical and efficient way to sample from the posterior by focusing on the most relevant parts of the hypothesis space.
In addition, since calculating the entropy $H\left(\mathbf{y}_{T}\right)$ entails to obtain the joint probability of the generated tokens $p(\mathbf{y}_{T}) = (\omega^{\mathbf{y}_{T}}_1, \ldots, \omega^{\mathbf{y}_{T}}_T)$, entropy-based method may only be applicable to white-box LLMs. 

\subsection{Entropy Approximation}\label{sec: est}
The generation of LLMs is generally free-form, which makes the uncertainty estimation for in-context learning still different from well-studied classification models that have specific labels. Specifically, not only may the LLM not always be able to return an expected answer, but the generated sequence may also consist of placeholder tokens. Calculating the entropy of the whole generated sequence would involve redundant tokens. Therefore, in this work, we propose to approximate the entropy of the output $H(\mathbf{y}_T)$, and the process is summarized in Figure \ref{fig: framework}. 

Given the probability distributions of the generated tokens $p(\mathbf{y}_{T})$ for one set of demonstrations, we only select token(s) $\omega^{\mathbf{y}_{T}}_t$ that directly answer the provided question. Taking the text classification task as an example, LLM is asked to directly output a numerical value standing for a predefined category (e.g., $0$: Sadness, $1$: Joy, etc.). The probability of the token $\omega^{\mathbf{y}_{T}}_t$ that represents the numerical value is then leveraged to denote the overall distribution of $p(\mathbf{y}_{T})$. We aggregate the answer probabilities from all $M$ decoded sequences and transform them as an answer distribution (as shown in the top right corner in Figure \ref{fig: framework}). After repeating the process $L$ times, where $L$ corresponds to $L$ different sets of demonstrations, we have a matrix $\mathcal{M}$ recording the answer distributions of choosing different demonstrations and model configurations (as shown in the lower right corner in Figure \ref{fig: framework}). The EU and AU can then be approximated as:
\begin{align*}
    EU &= \frac{1}{L}\sum H\left(\sigma(\mathcal{M}_{:,j})\right),\\
    AU &= H\left(\sigma\left( \sum[\mathcal{M}_{:,j}]\right) \right ) - \frac{1}{L}\sum H\left(\sigma\left(\mathcal{M}_{:,j}\right)\right),
\end{align*}
where $\sigma(\cdot)$ normalizes the column $\mathcal{M}_{:,j}$ into a probability distribution, and entropy $H(\cdot)$ can be calculated as $-\sum_{k=1}^{K}\left(p(\mathcal{M}_{k,j}) * \log(p(\mathcal{M}_{k,j}))\right)$ if the number of labels is $K$. Note that we have instructed LLMs to not generate tokens with less semantic meaning, such as dashes, spaces, or non-related words. In practice, our adopted LLMs can follow the instruction to only return desired answers so that the whole sentence will be the answer tokens (no need to select tokens).

\section{Related Works}
\paragraph{Uncertainty Quantification and Decomposition.} Uncertainty quantification aims to measure the confidence of models' predictions, which has drawn attention from various domains \cite{zhao2020uncertainty,ling2022source,Malo2014GoodDO}. Measuring uncertainty is essential in many real-world NLP applications where making a wrong prediction with high confidence can be disastrous (e.g., assessing the confidence in a translation or a generated piece of information). This is especially important in foundation models since we do not have enough resources to finetune the model \cite{abdar2021review}. To better understand the uncertainty, the primary focus is on understanding and categorizing the sources of uncertainty for interpreting the models' outputs more effectively. The output uncertainty can typically be categorized into \textit{Aleatoric Uncertainty} that arises from the inherent noise in the data, and \textit{Epistemic Uncertainty} that arises due to inappropriate model architecture or overfitted/underfitted parameters. Existing methods \cite{chowdhary2013distinguishing,depeweg2017uncertainty,malinin2020uncertainty} have come up with various methods (e.g., Bayesian neural network, Deep Ensembles, and Monte Carlo Dropout) to decompose the uncertainty.

\paragraph{Uncertainty in Language Models.} 
LLMs have revolutionized the learning and inference paradigm in many domains \citep{chen2024hytrel, chen2021explicitly}, but existing works using LLMs often neglect the importance of uncertainty in their responses. Earlier works \citep{xiao2019quantifying,desai2020calibration,jiang2021can} on uncertainty in language models have focused on the calibration of classifiers (e.g., applying dropout to the model parameters or leveraging ensemble voting) to better assess the confidence of the generated output. When it comes to the era of LLMs, multiple works \citep{xiao2021hallucination,xiao2022uncertainty,lin2022towards,yu2022efficient,lin2023generating,kuhn2023semantic,fadeeva2023lm} have been proposed to measure the uncertainty of LLM's prediction in multiple aspects (e.g., lexical uncertainty, text uncertainty, and semantic uncertainty) for multiple NLP tasks. Another line of works \cite{kadavath2022language,zhou2023navigating,amayuelas2023knowledge, chen2024hytrel} instead tries to analyze how to extract knowledge from a language model correctly and self-evaluate the correctness with a confidence score. 
However, despite these commendable efforts, existing methods still lack an effective way to directly quantify and decompose the uncertainty inherent in the outputs of LLMs with in-context learning.

\section{Experiments}
We evaluate the uncertainty decomposition procedure on realistic natural language understanding problems. By comparing state-of-the-art uncertainty quantification methods, we aim to examine what type of uncertainty is the most promising indicator of high confidence for LLMs. In addition, we also provide generalization analysis and two specific out-of-distribution detection applications. Due to the space limit, extra experiments and experiment settings are provided in the Appendix.

\subsection{Experiment Setup}
We evaluate the decomposed uncertainties on open-source LLMs with different model sizes. We leverage \textsc{LLaMA-2} \citep{touvron2023llama}, which is the most widely applied open LLM, with its $7$B, $13$B, and $70$B model sizes. The primary experiments are conducted with \textsc{LLaMA-2} models. In order to further demonstrate the generalization ability of our method, we apply our uncertainty quantification method on \textsc{OPT-13B} \citep{zhang2022opt}.

\noindent\textbf{Data.} We consider different Natural Language Understanding tasks. \textit{1) Sentiment Analysis}: \textsc{Emotion} \citep{saravia-etal-2018-carer} contains $2,000$ test cases and six classes; Financial Phrasebank (Financial) \citep{Malo2014GoodDO} contains $850$ financial news and three sentiment classes; Stanford Sentiment Treebank v2 (SST2) \citep{socher2013recursive} consists of $872$ sentences from movie reviews and two classes. \textit{2) Linguistic Acceptability:} The Corpus of Linguistic Acceptability (COLA) \citep{warstadt2019neural} is about English acceptability judgments, which has $1,040$ test cases and two classes. \textit{3) Topic Classification:} AG\_News \citep{Zhang2015CharacterlevelCN} contains $1,160$ test cases and four classes.

\begin{table*}[h]
\centering
\renewcommand{\arraystretch}{1.2} % Adjust vertical padding
\resizebox{0.92\textwidth}{!}{%
\begin{tabular}{@{}clcccccllcccc@{}}
\toprule
\multicolumn{1}{c}{\multirow{2}{*}{}} &
  \multicolumn{1}{c}{\multirow{2}{*}{\begin{tabular}[c]{@{}c@{}}Inference\\ Model\end{tabular}}} &
  \multirow{2}{*}{ACC} &
  \multicolumn{2}{c}{\textbf{Likelihood}} &
  \multicolumn{2}{c}{\textbf{Entropy}} &
  \multicolumn{2}{c}{\textbf{Semantic}} &
  \multicolumn{2}{c}{\cellcolor{red!15}\textbf{Ours (EU)}} &
  \multicolumn{2}{c}{\cellcolor{red!15}\textbf{Ours (AU)}} \\ \cmidrule(lr){4-5} \cmidrule(lr){6-7} \cmidrule(lr){8-9} \cmidrule(lr){10-11} \cmidrule(l){12-13}
\multicolumn{1}{c}{}     & \multicolumn{1}{c}{} &       & AUPR & ROC & AUPR & ROC & AUPR & ROC & AUPR & ROC & AUPR & ROC \\ \midrule
\multirow{6}{*}{\rotatebox[origin=c]{90}{\textbf{Emotion}}} & \textsc{LLaMA-7b-random}      & 0.407 & 0.423 & 0.426 & 0.448 & 0.501 & 0.598 & 0.607 & \textbf{0.688} & \textbf{0.667} & 0.625 & 0.579 \\
                         & \textsc{LLaMA-7b-class}       & 0.411 & 0.562 & 0.423 & 0.657 & 0.538 & 0.697 & 0.653 & \textbf{0.745} & \textbf{0.696} & 0.691 & 0.601 \\
                         & \textsc{LLaMA-13b-random}     & 0.501 & 0.597 & 0.613 & 0.584 & 0.503 & 0.612 & 0.625 & \textbf{0.645} & \textbf{0.681} & 0.559 & 0.585 \\
                         & \textsc{LLaMA-13b-class}      & 0.533 & 0.641 & 0.578 & 0.593 & 0.554 & 0.652 & 0.701 & \textbf{0.622} & \textbf{0.686} & 0.526 & 0.599 \\
                         & \textsc{LLaMA-70b-random}     & 0.584 & 0.512 & 0.462 & 0.491 & 0.452 & 0.657 & 0.696 & \textbf{0.667} & \textbf{0.713} & 0.531 & 0.663 \\
                         & \textsc{LLaMA-70b-class}      & 0.592 & 0.537 & 0.484 & 0.469 & 0.442 & 0.622 & 0.689 & \textbf{0.659} & \textbf{0.721} & 0.612 & 0.693 \\
\midrule
\multirow{6}{*}{\rotatebox[origin=c]{90}{\textbf{Financial}}} & \textsc{LLaMA-7b-random}  & 0.379 & 0.821 & 0.532 & 0.728 & 0.438 & 0.715 & 0.624 & \textbf{0.731} & \textbf{0.672} & 0.669 & 0.582 \\
                         & \textsc{LLaMA-7b-class}   & 0.397 & 0.593 & 0.505 & 0.548 & 0.362 & 0.732 & 0.699 & \textbf{0.803} & \textbf{0.711} & 0.753 & 0.589 \\
                         & \textsc{LLaMA-13b-random} & 0.476 & 0.894 & 0.571 & 0.652 & 0.463 & 0.705 & 0.545 & 0.718 & 0.512 & \textbf{0.729} & \textbf{0.573} \\
                         & \textsc{LLaMA-13b-class}      & 0.477 & 0.752 & 0.594 & 0.692 & 0.531 & 0.694 & 0.543 & \textbf{0.765} & \textbf{0.610} & 0.758 & 0.592 \\
                         & \textsc{LLaMA-70b-random}     & 0.530 & 0.816 & 0.509 & 0.754 & 0.493 & 0.679 & 0.688 & \textbf{0.779} & \textbf{0.754} & 0.734 & 0.642 \\
                         & \textsc{LLaMA-70b-class}      & 0.537 & 0.668 & 0.469 & 0.623 & 0.439 & 0.774 & 0.649 & \textbf{0.893} & \textbf{0.804} & 0.739 & 0.659 \\
\midrule
\multirow{6}{*}{\rotatebox[origin=c]{90}{\textbf{SST-2}}} & \textsc{LLaMA-7b-random}      & 0.856 & 0.149 & 0.636 & 0.135 & 0.587 & 0.244 & 0.593 & \textbf{0.286} & 0.683 & 0.205 & \textbf{0.702} \\
                         & \textsc{LLaMA-7b-class}       & 0.897 & 0.230 & 0.666 & 0.196 & 0.579 & 0.253 & 0.577 & 0.248 & \textbf{0.701} & \textbf{0.302} & 0.673 \\
                         & \textsc{LLaMA-13b-random}     & 0.866 & 0.268 & 0.472 & 0.204 & 0.467 & \textbf{0.355} & 0.712 & 0.314 & 0.677 & 0.326 & \textbf{0.816} \\
                         & \textsc{LLaMA-13b-class}      & 0.928 & 0.178 & 0.425 & 0.113 & 0.439 & 0.343 & 0.631 & \textbf{0.397} & \textbf{0.836} & 0.367 & 0.639 \\
                         & \textsc{LLaMA-70b-random}     & 0.932 & 0.091 & 0.597 & 0.137 & 0.475 & 0.258 & 0.565 & \textbf{0.318} & \textbf{0.764} & 0.298 & 0.571 \\
                         & \textsc{LLaMA-70b-class}      & 0.938 & 0.132 & 0.552 & 0.185 & 0.531 & 0.312 & 0.679 & 0.331 & \textbf{0.851} & \textbf{0.362} & 0.697 \\
\midrule
\multirow{6}{*}{\rotatebox[origin=c]{90}{\textbf{COLA}}} & \textsc{LLaMA-7b-random}      & 0.599 & 0.388 & 0.557 & 0.329 & 0.443 & 0.358 & 0.502 & \textbf{0.416} & \textbf{0.562} & 0.377 & 0.517 \\
                         & \textsc{LLaMA-7b-class}       & 0.639 & 0.392 & 0.523 & 0.381 & 0.478 & 0.425 & 0.526 & \textbf{0.473} & \textbf{0.587} & 0.401 & 0.506 \\
                         & \textsc{LLaMA-13b-random}     & 0.652 & 0.389 & 0.498 & 0.287 & 0.512 & 0.433 & 0.562 & 0.469 & \textbf{0.572} & \textbf{0.488} & 0.565 \\
                         & \textsc{LLaMA-13b-class}      & 0.649 & 0.412 & 0.418 & 0.342 & 0.517 & 0.426 & 0.548 & 0.456 & 0.568 & \textbf{0.523} & \textbf{0.641} \\
                         & \textsc{LLaMA-70b-random}     & 0.826 & 0.481 & 0.599 & 0.312 & 0.471 & 0.372 & 0.625 & 0.317 & \textbf{0.716} & \textbf{0.329} & 0.676 \\
                         & \textsc{LLaMA-70b-class}      & 0.852 & 0.357 & 0.612 & 0.397 & 0.588 & 0.397 & 0.613 & 0.389 & \textbf{0.727} & \textbf{0.425} & 0.682 \\
\midrule
\multirow{6}{*}{\rotatebox[origin=c]{90}{\textbf{AG\_News}}} & \textsc{LLaMA-7b-random}      & 0.646 & 0.238 & 0.472 & 0.265 & 0.463 & 0.312 & 0.612 & \textbf{0.448} & \textbf{0.634} & 0.361 & 0.537 \\
                         & \textsc{LLaMA-7b-class}       & 0.679 & 0.267 & 0.505 & 0.372 & 0.523 & 0.378 & 0.562 & \textbf{0.384} & \textbf{0.627} & 0.326 & 0.538 \\
                         & \textsc{LLaMA-13b-random}     & 0.685 & 0.365 & 0.517 & 0.364 & 0.522 & 0.374 & 0.548 & \textbf{0.395} & \textbf{0.648} & 0.378 & 0.552 \\
                         & \textsc{LLaMA-13b-class}      & 0.685 & 0.378 & 0.528 & 0.359 & 0.413 & 0.411 & 0.566 & \textbf{0.429} & \textbf{0.654} & 0.401 & 0.569 \\
                         & \textsc{LLaMA-70b-random}     & 0.792 & 0.311 & 0.478 & 0.316 & 0.498 & \textbf{0.401} & 0.552 & 0.309 & \textbf{0.635} & 0.319 & 0.543 \\
                         & \textsc{LLaMA-70b-class}      & 0.838 & \textbf{0.302} & 0.511 & 0.271 & 0.528 & 0.354 & 0.532 & 0.274 & \textbf{0.662} & 0.283 & 0.571 \\\bottomrule
\end{tabular}%
}
\vspace{-1mm}
\caption{The performance comparison on the misclassification rate based on the uncertainty score from each approach. For each dataset, correct predictions are labeled as $0$ and incorrect ones are labeled as $1$. We show the AUPR and ROC (the higher the better) based on the uncertainty score and misclassification rate with two types of demonstration selection strategy: \textsc{Random} and \textsc{Class} as well as three \textsc{LLaMA} model sizes: \textsc{7B}, \textsc{13B}, and \textsc{70B}.}
\vspace{-3mm}
\label{table: results}
\end{table*}

\noindent\textbf{Demonstration \& Model Configuration Sampling.} We evaluate each method on the testing set of each dataset and choose two strategies to randomly sample in-context learning demonstrations. 1) \textit{Random}: we randomly sample demonstrations (training instances with labels) from the training set regardless their labels. 2) \textit{Class}: we randomly sample demonstrations but ensure there is at least one demonstration per label class. To generate various sequences based on one set of demonstrations, we adopt Beam Search with beam width $=10$ to approximate the sampling process of $\Theta\sim q(\Theta)$.

\noindent\textbf{Comparison Methods.} Our study also evaluates the following baseline uncertainty estimation methods:
1) \textit{Likelihood-based Uncertainty} (Likelihood) \citep{malinin2020uncertainty} calculates the sum of log probabilities of all tokens generated from language models and normalizes it by the sequence length. 2) \textit{Entropy-based Uncertainty} (Entropy) \citep{xiao2019quantifying} calculates the entropy of the probability distributions of the generated tokens. 3) \textit{Semantic Uncertainty} (Semantic) \citep{kuhn2023semantic} is the most advanced entropy-based uncertainty estimation method, which groups generated sequences into clusters according to their semantic embeddings. The average entropy across all groups is viewed as the uncertainty score.

\noindent\textbf{Evaluation Metrics.} We show the prediction accuracy of each dataset. In addition, we leverage two standard metrics: the Area under Precision-Recall Curve (AUPR) and AUROC (ROC) to evaluate the uncertainty. AUPR calculates the area under the Precision-Recall curve. AP is high when both precision and recall are high, and low when either of them is low across a range of confidence thresholds. ROC represents the likelihood that a correct answer is selected. An ideal ROC rating is $1$, whereas a random uncertainty estimate would yield ROC $=0.5$.

% Due to space limit, details of hyper-parameter settings and actual in-context learning prompts for each dataset are provided in Appendix \ref{sec: appendix_exp}. 

% \begin{figure}%
%     \centering
%     \subfloat[\centering ROC Curve by \textsc{OPT-13B}]{{\includegraphics[width=0.4\textwidth]{figures/opt_roc_cropped.pdf}}}%
%     \vspace{-3mm}
%     \subfloat[\centering ROC Curve by \textsc{LLaMA-2-13B}]{{\includegraphics[width=0.4\textwidth]{figures/llama_roc_cropped.pdf} }}%
%     \caption{2 Figures side by side}%
%     \label{fig:generalization}%
%     \vspace{-7mm}
% \end{figure}

\subsection{Quantitative Analysis}
We compare the performance of different methods in assessing the misclassification samples based on their perspective uncertainty scores. We follow the procedure: 1) We use LLMs to classify all examples in the dataset with different beam search branches and demonstrations; 2) we use different uncertainty quantification methods to obtain a score associated with each test instance; 3) we assign each example a $0$ if it was classified correctly or a $1$ if it was misclassified; and 4) we calculate AUPR and AUROC based on the misclassification rate and uncertainty score. Ideally, misclassified samples should have higher uncertainty scores. The results are shown in Table \ref{table: results}. Note that our proposed method can decompose the uncertainty into epistemic uncertainty (EU) and aleatoric uncertainty (AU), we thus show the performance of EU and AU separately. 

As shown in the table, in most cases, our proposed methods (EU and AU) consistently show higher AUPR and ROC scores across all datasets, which indicates a better performance in assessing misclassification samples based on uncertainty scores. Moreover, we also draw some observations from the table. \textit{1. Class Sampling Strategy Proves Superior}: The class sampling strategy generally yields higher AUPR and ROC scores across datasets, which proves it is more effective than random demonstration sampling. Class sampling ensures that each class is represented in the sample and reduces sampling bias, which is crucial in scenarios where the dataset might be imbalanced or where certain classes are underrepresented. \textit{2) Increasing Model Size Enhances Performance}: Larger models (moving from 7B to 70B) tend to have better performance in terms of AUPR and ROC. Specifically, there's a general trend of increasing AUPR and ROC scores as model size increases from 7B to 13B to 70B for all comparison methods. Some datasets and metrics do not strictly follow this trend. For instance, in the \textsc{Emotion} dataset, the 70B model sometimes shows a slight decrease in performance compared to the 13B model. The inconsistencies in performance improvement with larger models, especially for EU, hint at the complexity of uncertainty assessment in different contexts and datasets. \textit{3. Treating all tokens equally can be harmful in uncertainty quantification}: both Likelihood and Entropy Uncertainty treat all tokens equally. However, some tokens carry greater relevance and representativeness than others, owing to the phenomenon of ``linguistic redundancy”. However, most uncertainty estimation methods treat all tokens with equal importance when estimating uncertainty, disregarding these inherent generative inequalities.

% Moreover, as can be seen from Table 1, EU only consistently performs better than AU in EMOTION (six classes) and AG\_News datasets (four classes), which are the most challenging datasets in our experiments. We provide our understanding here: To let LLMs make better predictions, we should use as many demonstrations as possible. However, the LLM’s context window size and the average length of test instances limit the number of demonstrations that can be used. In this case, different demonstrations may make fewer contributions to guide the model to make better performance, which makes AU less effective than EU in indicating uncertainties. For other less challenging datasets, e.g., SST2 and COLA, different demonstrations can indeed make the AU perform better than EU. In addition, the adopted LLMs (OPT-13B and LLaMA models) are already strong enough to make zero-shot predictions since they have likely observed the dataset during their retraining phase.

\begin{figure*}%
    \subfloat[\centering PR by \textsc{OPT-13B}]{{\includegraphics[width=0.25\textwidth]{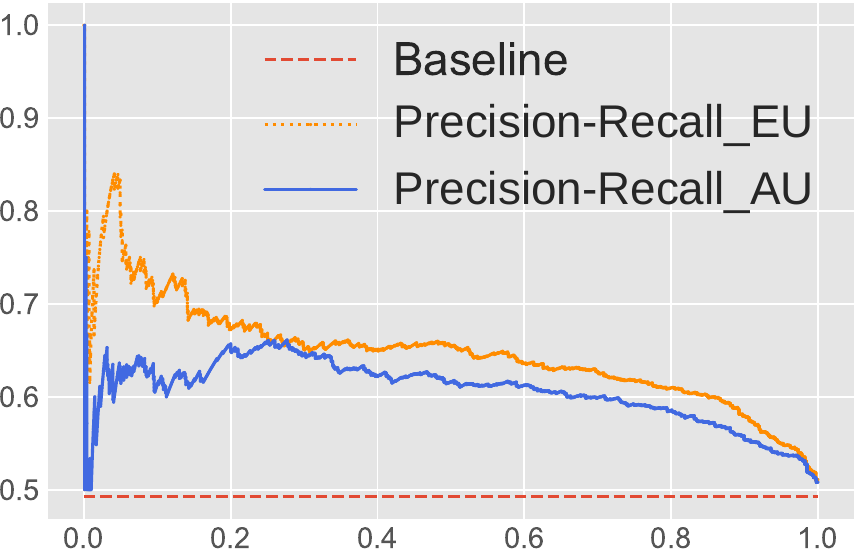}}}%
    \subfloat[\centering PR by \textsc{LLaMA-2-13B}]{{\includegraphics[width=0.25\textwidth]{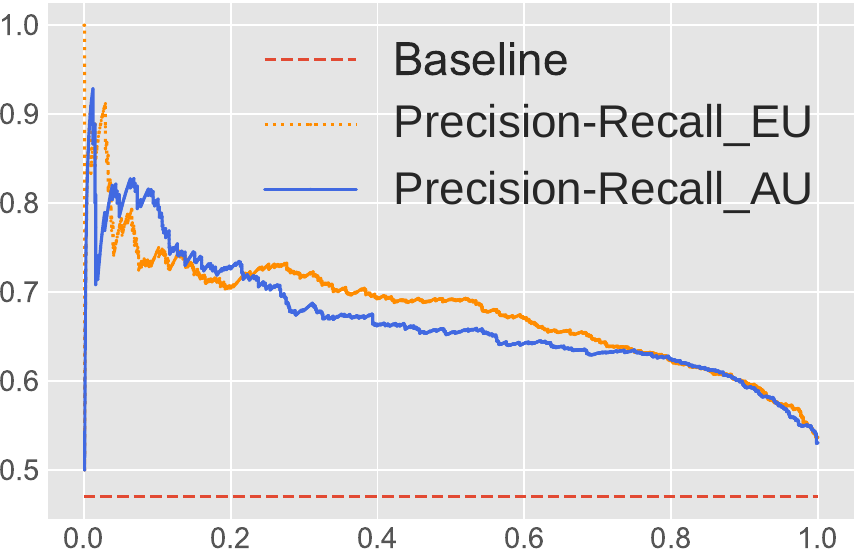}}}%
    \subfloat[\centering ROC by \textsc{OPT-13B}]{{\includegraphics[width=0.25\textwidth]{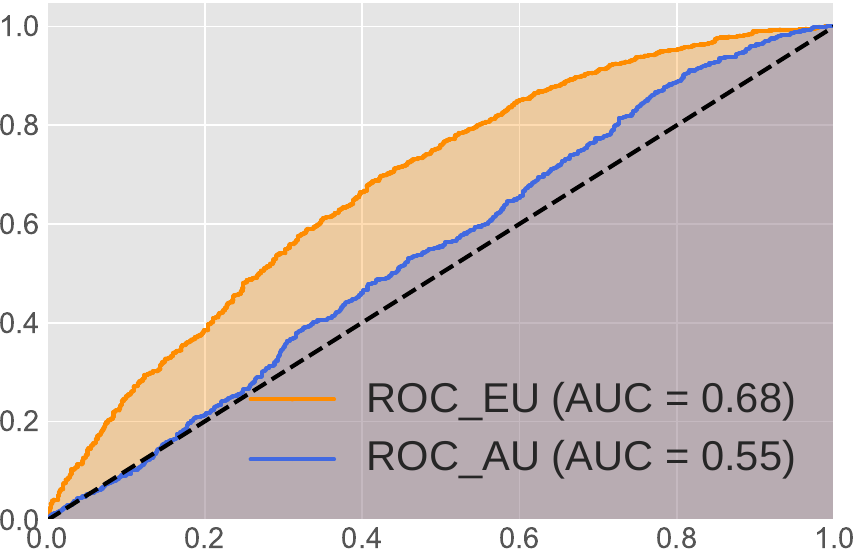} }}%
    \subfloat[\centering ROC by \textsc{LLaMA-2-13B}]{{\includegraphics[width=0.25\textwidth]{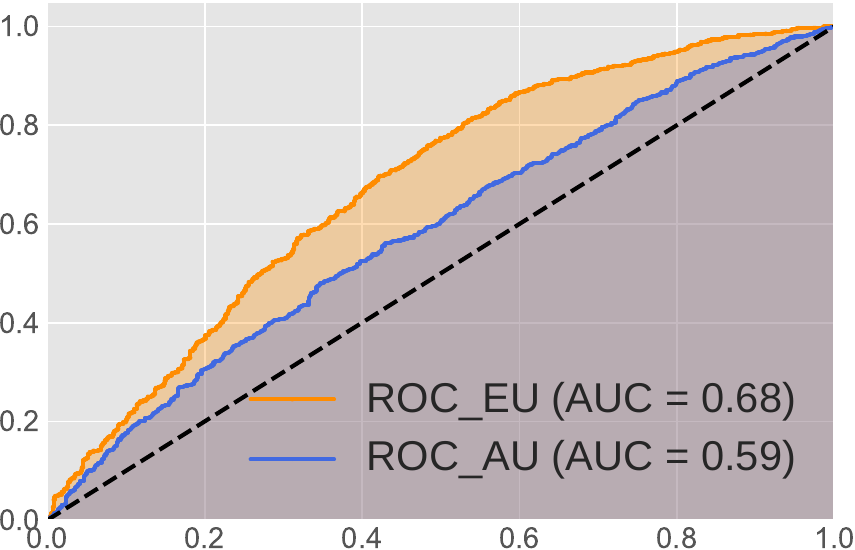} }}%
    \vspace{-3mm}
    \caption{The performance of misclassification rate using two backbone LLMs: \textsc{OPT-13B} and \textsc{LLaMA-2-13B} on \textsc{Emotion} dataset. (a) and (b) demonstrate the precision-recall curves (x-axis is the recall and y-axis is the precision) for \textsc{OPT-13B} and \textsc{LLaMA-2-13B}; (c) and (d) demonstrate the ROC curve  (x-axis is the false positive rate and y-axis is the true positive rate) for \textsc{OPT-13B} and \textsc{LLaMA-2-13B}, respectively.}%
    \label{fig:generalization}%
    \vspace{-5mm}
\end{figure*}

\subsection{Generalization Capability}
In this work, we also show how our method performs when applied to different LLMs. We compare the performance of misclassification rate when using \textsc{OPT-13B} and \textsc{LLaMA-2-13B}. We compute the precision-recall (PR) curve and ROC curve using two backbone LLMs on the \textsc{Emotion} dataset, and the results are shown in Figure \ref{fig:generalization}.

As shown in Figure \ref{fig:generalization}, our method exhibits consistent trends across different LLMs. The precision-recall curves of both uncertainties (Figure \ref{fig:generalization} (a) and \ref{fig:generalization} (b)) between the two methods are almost identical, and the model's capability between two LLMs is also reflected in the PR curves of EU. Furthermore, by comparing Figure \ref{fig:generalization} (c) and \ref{fig:generalization} (d), the ROC curves of both LLMs show a similar pattern, with the AUC scores not deviating significantly. Specifically, both \textsc{OPT-13B} and \textsc{LLaMA-2-13B} exhibit the same Area Under ROC (AUROC) curve $=0.68$ for AU. Since \textsc{LLaMA-2-13B} is a more powerful LLM than \textsc{OPT-13B}, our method can quantify that EU of \textsc{LLaMA-2-13B} (AUROC $=0.59$) is better than \textsc{OPT-13B} (AUROC $=0.55$). This finding further supports our method maintains its performance irrespective of the underlying model and its robust generalization capability. 

\subsection{Misclassification Rate with Out of Domain Demonstration}\label{sec: mis}
Out-of-domain in-context Demonstration refers to the test instance being coupled with less relevant or out-of-domain demonstrations, which the model may be misled and not handle the test instance reliably. In this work, we analyze the misclassification rate of out-of-domain Demonstration in the \textsc{Emotion} dataset (six-class sentiment analysis task) by providing LLMs with relevant demonstrations (sampled from Finance Phrasebank which is a three-class sentiment analysis task) and complete out-of-domain demonstrations (sampled from COLA which is a binary linguistic acceptability task). We conduct the task with two demonstration selection strategies, and the results are provided in Table \ref{tab:OOD}. 

\begin{table}[!h]
\centering
\resizebox{0.46\textwidth}{!}{%
\begin{tabular}{@{}lcccc@{}}
\toprule
                                                        & \multicolumn{2}{c}{LLaMA-13B-Random} & \multicolumn{2}{c}{LLaMA-13B-Class} \\ \cmidrule(lr){2-3} \cmidrule(lr){4-5}
\multirow{-2}{*}{}                                      & EU                & AU               & EU               & AU               \\ \midrule
\rowcolor[HTML]{E0DCDC} 
\begin{tabular}[c]{@{}l@{}}Original\\ Demo\end{tabular} & 0.681             & 0.585            & 0.686            & 0.599            \\ \midrule
\begin{tabular}[c]{@{}l@{}}Relevant\\ Demo\end{tabular} & \begin{tabular}[c]{@{}c@{}}0.688\\ (\textcolor{blue}{$+1.0$\%)}\end{tabular} & \begin{tabular}[c]{@{}c@{}}0.541\\ (\textcolor{red}{$-7.5$\%})\end{tabular} & \begin{tabular}[c]{@{}c@{}}0.671\\ (\textcolor{red}{$-2.2$\%})\end{tabular} & \begin{tabular}[c]{@{}c@{}}0.524\\ (\textcolor{red}{$-12.5$\%})\end{tabular} \\ \midrule
\begin{tabular}[c]{@{}l@{}}OOD\\ Demo\end{tabular}      &       \begin{tabular}[c]{@{}c@{}}0.671\\ (\textcolor{red}{$-1.4$\%})\end{tabular}            &        \begin{tabular}[c]{@{}c@{}}0.501\\ (\textcolor{red}{$-13.3$\%})\end{tabular}          &       \begin{tabular}[c]{@{}c@{}}0.673\\ (\textcolor{red}{$-1.8$\%})\end{tabular}           &        \begin{tabular}[c]{@{}c@{}}0.497\\ (\textcolor{red}{$-17.0$\%})\end{tabular}          \\ \bottomrule
\end{tabular}%
}
\caption{Comparison of AUROC in misclassificatin rate on \textsc{Emotion} dataset, where ``Original Demo'' indicates we sample demonstrations from its original training set, ``Relevant Demo'' indicates we sample demonstrations from {Finance Phrasebank} Dataset (a relevant sentiment analysis task, and ``OOD Demo'' indicates we sample demonstrations from an irrelevant dataset: \textbf{COLA}.}
\vspace{-3mm}
\label{tab:OOD}
\end{table}
As shown in the table, changes in the performance of the EU are relatively minor under all conditions, suggesting that the model is more stable or less sensitive to the changes in demonstration data within this metric. In contrast, the AU shows more significant fluctuations, which implies that the AU is more sensitive to the quality and relevance of demonstration data. When relevant demonstrations from the Finance Phrasebank sentiment analysis dataset are used, there's a slight improvement or a minor decrease in EU, but a notable decrease in AU. This suggests that even relevant but not identical data can confuse the model, especially for the AU. With out-of-domain demonstrations from COLA, the model's performance drops more significantly, with the AU metric showing a dramatic decrease of up to 17.0\%, which indicates that the model struggles significantly when the demonstrations are not relevant to the task it's being tested on.

\subsection{Out-of-domain Demonstration Detection}
Out-of-domain (OOD) demonstration refers to coupling a test instance with less relevant or OOD demonstrations, potentially leading the model to be misled and handle the test instance unreliably. In this study, we investigate whether uncertainty scores can effectively distinguish between in-domain and OOD demonstrations. In our labeling scheme, in-domain demonstrations are labeled as $0$, while OOD demonstrations are labeled as $1$. AUPR and ROC analyses are performed based on the labels and uncertainty scores, with results summarized in Table \ref{tab: out-of-domain}. Specifically, we conduct experiments on the \textsc{Emotion} dataset, involving two scenarios: in-domain demonstrations (sampled from its training set) and relevant demonstrations (sampled from Finance Phrasebank, a three-class sentiment analysis task). Additionally, we compare in-domain demonstrations with complete OOD demonstrations (sampled from COLA, a binary linguistic acceptability task).

\begin{table}[t]
\centering
\resizebox{0.47\textwidth}{!}{%
\begin{tabular}{@{}lllllll@{}}
\toprule
\multirow{2}{*}{}                                                           & \multicolumn{2}{c}{Semantic} & \multicolumn{2}{c}{Ours (EU)} & \multicolumn{2}{c}{Ours (AU)} \\ \cmidrule(lr){2-3}\cmidrule(lr){4-5}\cmidrule(lr){6-7} 
 & AUPR & ROC & AUPR & ROC & AUPR & ROC \\ \midrule
\multicolumn{1}{c}{\begin{tabular}[c]{@{}c@{}}Relevant\\ Demo\end{tabular}} &      0.702         &      0.644        &    \cellcolor[HTML]{DAE8FC}\textbf{0.742}         & \cellcolor[HTML]{DAE8FC}\textbf{0.935}           &      0.657       & 0.682           \\\midrule
\multicolumn{1}{c}{\begin{tabular}[c]{@{}c@{}}OOD\\ Demo\end{tabular}}      &       0.698        &     0.712         &       \cellcolor[HTML]{DAE8FC}\textbf{0.784}      & \cellcolor[HTML]{DAE8FC}\textbf{0.941}           &       0.773      & 0.607           \\ \bottomrule
\end{tabular}%
}
\vspace{-3mm}
\caption{Out-of-domain demonstration detection conducted with \textsc{LLAMA-2-13B} on \textsc{Emotion} Dataset.}
\label{tab: out-of-domain}
\vspace{-4mm}
\end{table}

As shown in Table \ref{tab: out-of-domain}, compared to the state-of-the-art Semantic Uncertainty and the AU, the EU demonstrates the best indicator to detect both less relevant and OOD demonstrations. Intuitively, the model's predictions would be impacted by irrelevant and OOD demonstrations and exhibit large variance. AU is less effective than EU in detecting OOD demonstrations since the demonstrations already have large inherent variability. Semantic Uncertainty instead cannot really distinguish what is the root cause of the predictive uncertainty.

\subsection{Semantic Out-of-distribution Detection}
Semantic out-of-distribution (SOOD) detection refers to distinguishing test samples with semantic shifts from the given demonstrations and the prompt. In this study, we mask out a few classes and ask LLMs to classify test samples into the rest of the classes. The method is expected to return a higher uncertainty score of SOOD test samples. Specifically, we mask two classes \textit{1: sadness} and \textit{2: anger} out of six classes from the \textsc{Emotion} dataset and ask LLM to categorize a given test sample only into the rest four classes. The SOOD samples are labeled as $1$ and in-distribution samples are labeled as $0$. Results of AUPR and ROC are recorded in Table \ref{tab: out-of-dis} in terms of different model sizes.

\begin{table}[!t]
\centering
\resizebox{0.47\textwidth}{!}{%
\begin{tabular}{@{}ccccccc@{}}
\toprule
\multirow{2}{*}{}                                                           & \multicolumn{2}{c}{Semantic} & \multicolumn{2}{c}{Ours (EU)} & \multicolumn{2}{c}{Ours (AU)} \\ \cmidrule(lr){2-3}\cmidrule(lr){4-5}\cmidrule(lr){6-7} 
 & AUPR & ROC & AUPR & ROC & AUPR & ROC \\ \midrule
\multicolumn{1}{c}{7B} &       0.477        &      0.532        &    \cellcolor[HTML]{DAE8FC}\textbf{0.548}         & \cellcolor[HTML]{DAE8FC}\textbf{0.658}           &      0.461       & 0.570           \\
13B      &       0.417        &      0.468        &      \cellcolor[HTML]{DAE8FC}\textbf{0.525}      &     \cellcolor[HTML]{DAE8FC}\textbf{0.592}       &      0.414       &      0.437      \\\bottomrule
\end{tabular}%
}
\vspace{-3mm}
\caption{Semantic out-of-distribution detection using \textsc{LLAMA-2} 7B and 13B on \textsc{Emotion} Dataset.}
\label{tab: out-of-dis}
\vspace{-7mm}
\end{table}

As shown in the table, EU still performs the best as a better indicator to recognize SOOD samples across different model sizes. SOOD samples are semantically different from the provided demonstrations, and the task description also masks out the correct class of these SOOD samples, leading to higher uncertainty in the model's predictions. Given the inappropriate task description and demonstrations, AU may not necessarily perform better in the presence of SOOD samples.

\section{Conclusion}
We provide a novel approach to decompose the predictive uncertainty of LLMs into its aleatoric and epistemic perspectives from the Bayesian perspective. We also design novel approximation methods to quantify different uncertainties based on the decomposition. Extensive experiments are conducted to verify the effectiveness and better performance of the proposed method than others. We believe this research stands as a significant stride toward harnessing the full potential of LLMs while being acutely aware of their performance boundaries. For future works, we plan to extend our method to other forms of data \citep{chen2022crossroads} and tasks \cite{zhang2024elad} to quantify the uncertainty.

\section*{Limitations}
The proposed work aims at quantifying predictive uncertainty and decomposing the value into its aleatoric and epistemic components. While we can achieve the best result compared to other methods, the proposed framework may only be applied in natural language understanding tasks (e.g., multiple-choice QA, text classification, linguistics acceptability, etc.). The proposed uncertainty estimation algorithm may have limited usage in quantifying uncertainties of generation tasks since we cannot tell which part of the generated sequence is semantically important.

\bibliography{anthology,custom}

\appendix

\section{Appendix}\label{sec:appendix}
\subsection{Variance-based Decomposition}\label{sec: var} Alternatively, we can use the variance as a measure
of uncertainty. Let $\sigma^2(\cdot)$ compute the variance of a probability distribution, and the total uncertainty present in Eq. (\ref{eq:BNN})
is then $\sigma^2(\mathbf{y}_{T}|\mathbf{x}_{1:T})$. This quantity can then be decomposed using the law of total variance:
\begin{align}\label{eq:variance}
    \sigma^2(\mathbf{y}_{T}|\mathbf{x}_{1:T})=&\sigma^2_{q(\Theta)}\left(\mathbb{E}[\mathbf{y}_{T}|\mathbf{x}_{1:T},\Theta]\right)\\&+\mathbb{E}_{q(\Theta)}\left[\sigma^2(\mathbf{y}_{T}|\mathbf{x}_{1:T},\Theta)\right].\nonumber
\end{align}
where $\mathbb{E}[\mathbf{y}_{T}|\mathbf{x}_{1:T},\Theta]$ and $\sigma^2(\mathbf{y}_{T}|\mathbf{x}_{1:T},\Theta)$ are mean and variance of $\mathbf{y}_{T}$ given $p\left(\mathbf{y}_{T}|\mathbf{x}_{1:T}, \Theta\right)$. $\sigma^2_{q(\Theta)}\left(\mathbb{E}[\mathbf{y}_{T}|\mathbf{x}_{1:T},\Theta]\right)$ represents the variance of $\mathbb{E}[\mathbf{y}_{T}|\mathbf{x}_{1:T},\Theta]$ when $\Theta \sim q(\Theta)$, which indicates the epistemic uncertainty since it ignores the contribution of $z$. In contrast, $\mathbb{E}_{q(\Theta)}\left[\sigma^2(\mathbf{y}_{T}|\mathbf{x}_{1:T},\Theta)\right]$ in Eq. (\ref{eq:variance}) represents the aleatoric uncertainty since it denotes the average value of $\sigma^2(\mathbf{y}_{T}|\mathbf{x}_{1:T},\Theta)$ with $\Theta \sim p(\Theta)$ and ingores the contribution of $\Theta$ to $\mathbf{y}_{T}$. %In practice, the Aleatoric and Epistemic Uncertainty of variance based decomposition can be calculated as:
% \begin{align}
%     \text{Aleatoric Uncertainty:} \\
%     \text{Epistemic Uncertainty:}
% \end{align}
Note that variance-based uncertainty decomposition does not involve the probability of the generated tokens, which is applicable to black-box LLMs (e.g., GPT models).

\paragraph{Variance Approximation.} In practice, when we are dealing with black-box LLMs (e.g., ChatGPT), there are multiple hyperparameters (e.g., \texttt{temperature} and \texttt{top\_p}) allowing to return different responses. Specifically, $[\mathbf{y}^{1}_{T}, \ldots, \mathbf{y}^{L}_{T}]$ can be obtained through querying the LLM with different demonstrations $[\mathbf{x}^{1}_{1:T-1}, \ldots, \mathbf{x}^{L}_{1:T-1}]$ $L$ times.
The different set of parameter configurations are denoted as $[\Theta_1, \ldots, \Theta_M]$. 
The $\mathbb{E}[\mathbf{y}_{T}|\mathbf{x}_{1:T},\Theta]$ can then be calculated as the expected model output given the input data and the model parameters $\Theta$. Calculate the variance of this expectation with respect to a set of model configurations over all sets of demonstrations gives the epistemic uncertainty. The variance $\sigma^2(\mathbf{y}_T)$ can also be obtained given a set of few-shot demonstrations over all model parameters. Finally, average this variance over the certain model configuration to obtain the aleatoric uncertainty. 

\subsection{Dataset Description}\label{sec: data}
\noindent\textbf{Sentiment Analysis.} 1) \textsc{Emotion} \citep{saravia-etal-2018-carer} contains $2,000$ test cases, where LLMs are asked to classify a given sentence with six categories: \textit{sadness}, \textit{joy}, \textit{love}, \textit{anger}, \textit{fear}, \textit{surprise}. 2) Financial Phrasebank (Financial) \citep{Malo2014GoodDO} contains $850$ test cases, where LLMs are asked to classify a given financial news with three categories: \textit{negative}, \textit{neutral}, \textit{positive}. 3) Stanford Sentiment Treebank v2 (SST2) \citep{socher2013recursive} consists of $872$ sentences from movie reviews and human annotations of their sentiment, where the language model is asked to predict the sentiment from two classes: \textit{positive} and \textit{negative}.

\noindent\textbf{Linguistic Acceptability.} 1) The Corpus of Linguistic Acceptability (COLA) \citep{warstadt2019neural} is about English acceptability judgments drawn from books and journal articles on linguistic theory. Each example is a sequence of words annotated with whether it is a grammatical English sentence, and there are $1,040$ test cases in total.

\noindent\textbf{Topic Classification.} TC aims at categorizing the given sentence into predefined topics. We adopt {AG\_News} \citep{Zhang2015CharacterlevelCN} is a dataset that collects more than $1$ million news articles, where LLMs are asked to classify a given news into four categories: \textit{World}, \textit{Sports}, \textit{Business}, and \textit{Sci/Tech}. There are $1,160$ test cases in total.

\subsection{Experiment Setup}\label{sec: appendix_exp}
We conduct experiments primarily on \textsc{llama-2-7b-chat-hf}, \textsc{llama-2-13b-chat-hf}, and \textsc{llama-2-70b-chat-hf}, where the model weights are downloaded from the website\footnote{https://ai.meta.com/resources/models-and-libraries/llama-downloads/}. Since we cannot actually ``sample'' model weights as Bayesian Neural Networks, in order to let LLMs return different outputs, we leverage Beam Search since it considers multiple best options based on beam width using conditional probability, which is better than the sub-optimal Greedy search. The beam search is conducted with the beam size $10$ and the max number of new tokens is set to be $16$ uniformly across all datasets. We choose a different number of demonstrations (details are recorded in Table \ref{tab: demonstrations}) to allow the LLM to achieve the best performance on each dataset, and we sample demonstrations four times uniformly across all datasets. 

\begin{table}[h]
\centering
\resizebox{0.6\columnwidth}{!}{%
\begin{tabular}{@{}lcc@{}}
\toprule
          & Random & Class       \\ \midrule
Emotion   & 6      & 1 per class \\
Financial & 6      & 2 per class \\
SST2      & 4      & 2 per class \\
COLA      & 2      & 1 per class \\
AG\_News  & 4      & 1 per class \\ \bottomrule
\end{tabular}%
}
\caption{Number of demonstrations selected in each dataset.}
\label{tab: demonstrations}
\end{table}

\begin{table*}[t]
\centering
\resizebox{\textwidth}{!}{%
\begin{tabular}{@{}ll@{}}
\toprule
\textbf{System Prompt} &
  \begin{tabular}[c]{@{}l@{}}\#\#\# Below is an instruction that describes a task. Clearly follow the instruction and write a short \\ response to answer it.\end{tabular} \\ \midrule
\textbf{Task Description} &
  \begin{tabular}[c]{@{}l@{}}\#\#\# Instruction: Classify the sentiment in the following text based on the six categories: \\ {[}0: Sadness; 1: Joy, 2: Love; 3: Anger; 4: Fear, 5: Surprise{]}. Provide the information in a \\ structured format WITHOUT additional comments, I just want the numerical label for each text.\end{tabular} \\ \midrule
\textbf{Demonstrations} &
  \begin{tabular}[c]{@{}l@{}}\#\#\# Here are some examples:\\ Example 1: Sentence: \{i didnt feel humiliated\} Category: \{0: Sadness\}\\ Example 2: Sentence: \{im grabbing a minute to post i feel greedy wrong\} Category: \{3: anger\}\\ Example 3: Sentence: \{i have the feeling she was amused and delighted\} Category: \{1: joy\}\\Example 4: Sentence: \{i feel more superior dead chicken or grieving child\} Category: \{1: joy\}\\Example 5: Sentence: \{i get giddy over feeling elegant in a pencil skirt\} Category: \{1: joy\}\\ ...\end{tabular} \\ \midrule
\textbf{Test Query} &
  \begin{tabular}[c]{@{}l@{}}\#\#\# Test\\ Sentence: \{\} Category: \end{tabular} \\ \bottomrule
\end{tabular}%
}

\caption{Prompt Template consists of four parts: 1) \textit{System Prompt} aims at providing a basic hint of the task; 2) \textit{Task Description} provides some details of the task, e.g., if it is a sentiment analysis task or how many labels are there; 3) \textit{Few-shot Demonstrations} are leveraged to give LLMs some basic formats of how the returned responses can be constructed; and 4) \textit{Test Query} is the final test query that we want LLMs to classify/categorize, and the LLM is only expected to return an exact answer to solve the given question.}
\label{tab: prompts}
\end{table*}

\subsection{Prompt Template}
In this work, we uniformly apply the following prompt template for all datasets. Take the \textsc{Emotion} dataset as an example, we summarize the prompt in Table \ref{tab: prompts}. Note that all datasets use the same template, small modifications are made on the actual label information and different demonstration numbers of different datasets.

\begin{table*}[t]
\centering
\resizebox{0.9\textwidth}{!}{%
\begin{tabular}{@{}llccc@{}}
\toprule
\multicolumn{2}{l}{\begin{tabular}[c]{@{}l@{}}\textbf{Testing Query}: \\ I had stated to her the reason I feel so fearful is because I feel unsafe (4: fear)\end{tabular}} &
  \begin{tabular}[c]{@{}c@{}}Extracted \\ Predictions\end{tabular} &
  EU &
  AU \\ \midrule
\multicolumn{1}{l|}{\multirow{7}{*}{LLaMA-2-7B}} &
  \begin{tabular}[c]{@{}l@{}}1. i felt anger when at the end of a telephone call (3: anger)\\ 2. i feel a little mellow today (1: joy)\\ 3. i don t feel particularly agitated (4: fear)\\ 4. i hate it when i feel fearful for absolutely no reason (4: fear)\\ 5. im updating my blog because i feel shitty (0: sadness)\end{tabular} &
  \begin{tabular}[c]{@{}c@{}}0, 0, 0, 1, 3\\ 4, 3, 4, 4, 4\end{tabular} &
  0.171 &
  0.372 \\ \cmidrule(l){2-5} 
\multicolumn{1}{l|}{} &
  \begin{tabular}[c]{@{}l@{}}1. i am feeling outraged it shows everywhere (4: fear)\\ 2. i do feel insecure sometimes but who doesnt (4: fear)\\ 3. i start to feel emotional (0: sadness)\\ 4. i feel so cold a href http irish (3: anger)\\ 5. i feel i have to agree with her even though i can imagine\\      some rather unpleasant possible cases (0: sadness)\end{tabular} &
  \begin{tabular}[c]{@{}c@{}}4, 4, 1, 3, 4\\ 4, 4, 4, 5, 4\end{tabular} &
  0.163 &
  0.189 \\ \midrule
\multicolumn{1}{l|}{\multirow{7}{*}{LLaMA-2-70B}} &
  \begin{tabular}[c]{@{}l@{}}1. i felt anger when at the end of a telephone call (3: anger)\\ 2. i feel a little mellow today (1: joy)\\ 3. i don t feel particularly agitated (4: fear)\\ 4. i hate it when i feel fearful for absolutely no reason (4: fear)\\ 5. im updating my blog because i feel shitty (0: sadness)\end{tabular} &
  \begin{tabular}[c]{@{}c@{}}4, 3, 4, 3, 4\\ 4, 4, 2, 4, 4\end{tabular} &
  0.012 &
  0.079 \\ \cmidrule(l){2-5} 
\multicolumn{1}{l|}{} &
  \begin{tabular}[c]{@{}l@{}}1. i am feeling outraged it shows everywhere (4: fear)\\ 2. i do feel insecure sometimes but who doesnt (4: fear)\\ 3. i start to feel emotional (0: sadness)\\ 4. i feel so cold a href http irish (3: anger)\\ 5. i feel i have to agree with her even though i can imagine\\      some rather unpleasant possible cases (0: sadness)\end{tabular} &
  \begin{tabular}[c]{@{}c@{}}4, 4, 4, 4, 4\\ 4, 4, 4, 4, 4\end{tabular} &
  0.004 &
  0.009 \\ \bottomrule
\end{tabular}%
}
\caption{Case study on the actual EU and AU decomposed from the predictive uncertainty}
\label{tab: case_study}
\end{table*}

\subsection{Case Study} 
Table \ref{tab: case_study} demonstrates the actual changes in AU and EU when presenting LLMs with different sizes and different demonstrations. Given the test query is: \textit{I had stated to her the reason I feel so fearful is that I feel unsafe} with the ground truth label is  \textit{(4: fear)}, which is a sentence with a negative feeling. For \textsc{LLaMA-2-7B}, by presenting LLMs with more diverse demonstrations (containing both positive and negative sentences), the results would be more diverse between different beam search returned sequences, leading to a relatively higher AU than EU.  For \textsc{LLaMA-2-70B} with a larger parameter space and model capability, the EU and AU are significantly reduced, which indicates the model is more confident in the generated output and the variation of data may not influence much to the prediction.

\end{document}